# An Automatic Clustering Technique for Optimal Clusters

[1]K. Karteeka Pavan, [2]Allam Appa Rao, [3]A.V. Dattatreya Rao
[1]Department of Computer Applications, Rayapati Venkata Ranga Rao and Jagarlamudi Chadramouli College of Engineering, Guntur, India
[2]Jawaharlal Nehru Technological University, Kakinada, India
[3]Department of Statistics, Acharya Nagarjuna University, Guntur, India
[1]karteeka@yahoo.com, [2]apparaoallam@gmail.com, [3]avdrao@gmail.com

**Abstract**- This paper proposes a simple, automatic and efficient clustering algorithm, namely, Automatic Merging for Optimal Clusters (AMOC) which aims to generate nearly optimal clusters for the given datasets automatically. The AMOC is an extension to standard k-means with a two phase iterative procedure combining certain validation techniques in order to find optimal clusters with automation of merging of clusters. Experiments on both synthetic and real data have proved that the proposed algorithm finds nearly optimal clustering structures in terms of number of clusters, compactness and separation.

Keywords : Clustering, Optimal clusters, k-means, validation technique

**1 Introduction**

The two fundamental questions in data clustering are to find number of clusters and their compositions. There are many clustering algorithms to answer the latter problem, but not many methods for the former problem. Although a number of clustering methods have been proposed for the latter problem, they are facing the difficulties in meeting the requirements of automation, quality, simplicity and efficiency. Discovering an optimal number of clusters in a large data set is usually a challenging task. Cheung [20] studied a rival penalized competitive learning algorithm [9 -10] that has demonstrated a very good result in finding the cluster number. The algorithm is formulated by learning the parameters of a mixture model through the maximization of a weighted likelihood function. In the learning process, some initial seed centers move to the genuine positions of the cluster centers in a data set, and other redundant seed points will stay at the boundaries or outside of the clusters. Bayesian-Kullback Ying-Yang proposed a unified algorithm for both unsupervised and supervised learning [13], which provides a reference for solving the problem of selection of the cluster number. Lee and Antonsson [2] used an evolutionary method to dynamically cluster a data set. Sarkar,et al. [11] and Fogel et al. [8] are proposed an approach to dynamically cluster a data set using evolutionary programming, where two fitness functions are simultaneously optimized: one gives the optimal number of clusters, whereas the other leads to a proper identification of each cluster's centroid. Recently Swagatam Das and Ajith Abraham [18] proposed an

Automatic Clustering using Differential Evolution (ACDE) algorithm by introducing a new chromosome representation and Jain [1] explained few more methods to select k, the number of clusters. The majority of these methods to determine the best number of clusters may not work very well in practice. The clustering algorithms are required to be run several times for good solution, and model-based methods, such as cross-validation and penalized likelihood estimation, are computationally expensive.

This paper proposes a simple, automatic and efficient clustering algorithm, namely, Automatic Merging for Optimal Clusters (AMOC) which aims to generate nearly optimal clusters for the given datasets automatically. The AMOC is an extension to standard k-means, which combines the validation techniques into the clustering process so that high quality clustering results can be produced. The technique is a two-phase iterative procedure. In the first phase it produces clusters for a large k. In the second phase, iteratively a low probability cluster is merged with its closest cluster using a validation technique. Experiments on both synthetic and real data sets from UCI prove that the proposed algorithm finds nearly optimal results in terms of compactness and separation.

Section (2) deals with formulation of the proposed algorithm, while section (3) illustrates the effectiveness of the new algorithm experimenting results on synthetic, real, and micro array data sets. Finally concluding remarks are included in section (4).

## 2. Automatic Merging for Optimal Clusters (AMOC)

Let $P = \{P_1, P_2,\ldots, P_m\}$ be a set of m objects in which each object $P_i$ is represented as $[p_{i,1}, p_{i,2},\ldots p_{i,n}]$ where n is the number of features. The algorithm accepts large $k_{max}$ as the upper bound of the number of clusters and is taken to be $\sqrt{m}$ by intuition [12]. It iteratively merges the lower probability cluster with its closest cluster according to average linkage and validates the merging result using Rand Index.

**Steps:**

1. Initialize $k_{max} = \sqrt{m}$
2. Assign $k_{max}$ objects randomly to the cluster centroids
3. Find the clusters using k-means
4. Compute Rand index

5. Find a cluster that has least probability and merge with its closest cluster. Recompute centroids, Rand index and decrement the number of clusters by one. If the newly computed Rand index is greater than the previous Rand index, then update Rand Index, number of clusters and cluster centroids with the newly computed values.
6. If step 5 has been executed for each and every cluster, then go to step7, otherwise got to step5.
7. If there is a change in number of clusters, then go to step2, otherwise stop.

**3. Experimental Results**

To evaluate the performance of AMOC, we have tested it using both simulated and real data. The clustering results of AMOC are compared with these of k-means, fuzzy-kmeans, and Automatic clustering using Differential Evolution (ACDE) that determines optimal clusters automatically. The results are validated with the Rand, Adjusted Rand, DB, CS and Silhouette cluster validity measures and by identifying error rate using number of misclassifications.

In this AMOC the choice of initial centroids were selected at random and also done as suggested by Arthu and Vassilvitskii [4]. The performance of the algorithm is also compared with k-means++ [4].

The k-means and Fuzzy-kmeans algorithms are implemented with the number of clusters as equal to the number of classes in the ground truth.

**3.1 Experimental Data**

The efficiency of new algorithms are evaluated by conducting experiments on five artificial data sets, three real datasets down loaded from the web site UCI and two microarray data sets (two yeast data sets) downloaded from http://www.cs. washington.edu/homes/kayee/cluster [7].

The real data sets used:
1. Iris plants database (m = 150, n = 4, K = 3)
2. Glass (m = 214, n = 9, K = 6)
3. Wine (m = 178, n = 13, K = 3)

The real microarray data sets used:

1. The yeast cell cycle data [15] showed the fluctuation of expression levels of approximately 6000 genes over two cell cycles (17 time points). We used two different subsets of this data with independent external criteria. The first subset (the 5-phase criterion) consists of 384 genes whose expression levels peak at different time points corresponding to the five phases of cell cycle [15]. We expect clustering results to approximate this five class partition. Hence, we used the 384 genes with the 5- phase criterion as one of our data sets.
2. The second subset (the MIPS criterion) consists of 237 genes corresponding to four categories in the MIPS database [6]. The four categories (DNA synthesis and replication, organization of centrosome, nitrogen and sulphur metabolism, and ribosomal proteins) were shown to be reflected in clusters from the yeast cell cycle data [16].

The five synthetic data sets from $N_p(\mu, \Sigma)$ with specified mean vector and variance covariance matrix are as follows.

1. Number of elements, m=350, number of attributes, n=3, number of clusters, k=2 with
$$\mu 1 = \begin{pmatrix} 2 \\ 3 \\ 4 \end{pmatrix} \Sigma 1 = \begin{pmatrix} 1 & 0.5 & 0.3333 \\ & 1 & 0.6667 \\ & & 1 \end{pmatrix} \mu 2 = \begin{pmatrix} 7 \\ 6 \\ 9 \end{pmatrix} \Sigma 2 = \begin{pmatrix} 1 & 1 & 1 \\ & 2 & 2 \\ & & 3 \end{pmatrix}$$

2. The data set with m=400, n=3, k=4 and with
$$\mu 1 = \begin{pmatrix} -1 \\ -1 \end{pmatrix} \Sigma 1 = \begin{pmatrix} 0.65 & 0 \\ 0 & 0.65 \end{pmatrix} \mu 2 = \begin{pmatrix} 2 \\ 2 \end{pmatrix} \Sigma 2 = \begin{pmatrix} 1 & 0.7 \\ & 1 \end{pmatrix} \mu 3 = \begin{pmatrix} -3 \\ +3 \end{pmatrix} \Sigma 3 = \begin{pmatrix} 0.78 & 0 \\ 0 & 0.78 \end{pmatrix} \mu 4 = \begin{pmatrix} -6 \\ +4 \end{pmatrix} \Sigma 4 = \begin{pmatrix} 0.5 & 0 \\ 0 & 0.5 \end{pmatrix}$$

3. The data set m=300, n=2, k=3 and with
$$\mu 1 = \begin{pmatrix} -1 \\ -1 \end{pmatrix} \Sigma 1 = \begin{pmatrix} 1 & 0 \\ 0 & 1 \end{pmatrix} \mu 2 = \begin{pmatrix} 2 \\ 2 \end{pmatrix} \Sigma 2 = \begin{pmatrix} 1 & 0 \\ 0 & 1 \end{pmatrix} \mu 3 = \begin{pmatrix} -3 \\ +3 \end{pmatrix} \Sigma 3 = \begin{pmatrix} 0.7 & 0 \\ 0 & 0.7 \end{pmatrix}$$

4. The data set m=800, n=2, k=6 and with
$$\mu 1 = \begin{pmatrix} -1 \\ -1 \end{pmatrix} \Sigma 1 = \begin{pmatrix} 0.65 & 0 \\ 0 & 0.65 \end{pmatrix} \mu 2 = \begin{pmatrix} -8 \\ -6 \end{pmatrix} \Sigma 2 = \begin{pmatrix} 1 & 0.7 \\ & 1 \end{pmatrix} \mu 3 = \begin{pmatrix} -3 \\ +6 \end{pmatrix} \Sigma 3 = \begin{pmatrix} 0.2 & 0 \\ 0 & 0.2 \end{pmatrix} \mu 4 = \begin{pmatrix} -8 \\ +14 \end{pmatrix} \Sigma 4 = \begin{pmatrix} 0.5 & 0 \\ 0 & 0.5 \end{pmatrix}$$
$$\mu 5 = \begin{pmatrix} 10 \\ 12 \end{pmatrix} \Sigma 5 = \begin{pmatrix} 0.3 & 0 \\ 0 & 0.3 \end{pmatrix} \mu 6 = \begin{pmatrix} +14 \\ -14 \end{pmatrix} \Sigma 6 = \begin{pmatrix} 0.1 & 0 \\ 0 & 0.1 \end{pmatrix}$$

5. The data set m=180, n=8, k=3 and with

$$\mu_1 = \begin{pmatrix} 1 \\ 1 \\ 2 \\ 1 \\ 0.5 \\ 2 \\ 1 \\ 0.5 \end{pmatrix} \quad \Sigma_1 = \begin{pmatrix} 1 & 0.5 & 0.333 & 0.25 & 0.2 & 0.1667 & 0.1429 & 0.125 \\ & 1 & 0.667 & 0.5 & 0.4 & 0.3333 & 0.2857 & 0.25 \\ & & 1 & 0.75 & 0.6 & 0.5 & 0.4286 & 0.375 \\ & & & 1 & 0.8 & 0.6667 & 0.5714 & 0.5 \\ & & & & 1 & 0.8333 & 0.7143 & 0.625 \\ & & & & & 1 & 0.8571 & 0.75 \\ & & & & & & 1 & 0.875 \\ & & & & & & & 1 \end{pmatrix} \quad \mu_2 = \begin{pmatrix} 1 \\ 1 \\ 1 \\ 1 \\ 1 \\ 1 \\ 1 \\ 1 \end{pmatrix}$$

$$\Sigma_2 = \begin{pmatrix} 1 & 1 & 1 & 1 & 1 & 1 & 1 \\ & 2 & 2 & 2 & 2 & 2 & 2 \\ & & 3 & 3 & 3 & 3 & 3 \\ & & & 4 & 4 & 4 & 4 \\ & & & & 5 & 5 & 5 \\ & & & & & 6 & 6 \\ & & & & & & 7 & 7 \\ & & & & & & & 8 \end{pmatrix} \quad \mu_3 = \begin{pmatrix} 1 \\ -2 \\ 0 \\ -1 \\ 0 \\ -1 \\ -2 \\ -2 \end{pmatrix} \quad \Sigma_3 = \begin{pmatrix} 1 & -1 & -1 & -1 & -1 & -1 & -1 & -1 \\ & 2 & 0 & 0 & 0 & 0 & 0 & 0 \\ & & 3 & 1 & 1 & 1 & 1 & 1 \\ & & & 4 & 2 & 2 & 2 & 2 \\ & & & & 5 & 3 & 3 & 3 \\ & & & & & 6 & 4 & 4 \\ & & & & & & 7 & 5 \\ & & & & & & & 8 \end{pmatrix}$$

**3.2 Presentation of Results**

In this paper, while comparing the performance of AMOC with the other techniques we are concentrating on two major issues: 1) quality of the solution as determined by Error rate and cluster validity measures Rand, Adjusted Rand, DB, CS and Silhouette, 2) ability to find the optimal number of clusters. Since all the algorithms produce different results in different individual runs, we have taken 40 independent runs of each algorithm. The Rand [19], Adjusted Rand, DB [5], CS [3] and Silhouette [14] metrics values and the overall error rate of the mean-of-run solutions provided by the algorithms over the 10 datasets have been provided in Table-1. The table also shows the mean number of classes determined by each algorithm except k-means and fuzzy-k. All the results presented in this table are averages over 40 independent runs of each algorithm. The minimum and maximum error rates those found in 40 independent runs of each algorithm on each dataset are also tabulated in Table-1.

The above observations are presented graphically in the following figures. Figure1. to Figure2 represent the number of clusters identified by AMOC and ACDE in 40 independent runs. The figures demonstrate that AMOC is performing well when compared to ACDE in determining the clusters. Figure3 represent error rates obtained in 40 independent runs by AMOC and ACDE. Figures 4 to Figure5 are the clusters and their centroids obtained during the execution of the AMOC, in each iteration when choice of the initial k is 9.

Table-1:Validity measures along with error rates

| Dataset | Algorithm | No. of clusters, k | | Mean values of Cluster Validity Measures | | | | | Error rate | | |
|---|---|---|---|---|---|---|---|---|---|---|---|
| | | i/p k | o/p k | ARI | RI | SIL | DB | CS | Mean | Least | Maximum |
| Synthetic1 | k-means | 2 | 2 | 0.92 | 0.96 | 0.839 | 0.467 | 0.645 | 0.236 | 1.714 | 2.286 |
| | k-means++ | | | 0.925 | 0.962 | 0.839 | 0.466 | 0.567 | 1.914 | 1.714 | 2.286 |
| | fuzk | | | 0.899 | 0.95 | 0.839 | 0.468 | 0.52 | 2.571 | 2.571 | 2.571 |
| | AMOC(rand) | 19 | 2 | 0.92 | 0.96 | 0.839 | 0.467 | 0.749 | 2.029 | 1.714 | 2.286 |
| | AMOC(kmpp) | | 2 | 0.925 | 0.963 | 0.839 | 0.466 | 0.749 | 1.905 | 1.714 | 2.286 |
| | ACDE | | 3.05 | 0.85 | 0.925 | 0.643 | 0.772 | 1.348 | 51.56 | 0 | 96 |
| Synthetic2 | k-means | 4 | 4 | 0.821 | 0.927 | 0.718 | 0.58 | 1.178 | 19.1 | 2.4 | 67 |
| | k-means++ | | | 0.883 | 0.953 | 0.776 | 0.519 | 1.21 | 7.16 | 2.4 | 59.8 |
| | fuzk | | | 0.944 | 0.979 | 0.791 | 0.484 | 0.931 | 2.2 | 2.2 | 2.2 |
| | AMOC(rand) | 22 | 3.05 | 0.694 | 0.867 | 0.738 | 0.559 | 1.067 | 46.5 | 2.4 | 80.2 |
| | AMOC(kmpp) | | 3.8 | 0.885 | 0.953 | 0.788 | 0.499 | 0.946 | 8.79 | 2.4 | 34.4 |
| | ACDE | | 5.35 | 0.885 | 0.957 | 0.68 | 0.674 | 1.321 | 58.89 | 2.4 | 96.2 |
| Synthetic3 | k-means | 3 | 3 | 0.957 | 0.98 | 0.813 | 0.509 | 0.87 | 2.242 | 1 | 1 |
| | k-means++ | | | 0.97 | 0.987 | 0.823 | 0.761 | 0.92 | 1 | 1 | 50.67 |
| | fuzk | | | 0.97 | 0.987 | 0.823 | 0.5 | 0.96 | 1 | 1 | 1 |
| | AMOC(rand) | 17 | 2.9 | 0.93 | 0.966 | 0.805 | 0.52 | 0.791 | 7.6 | 1 | 67 |
| | AMOC(kmpp) | | 2.95 | 0.95 | 0.976 | 0.814 | 0.504 | 0.78 | 4.317 | 1 | 67.33 |
| | ACDE | | 4 | 0.472 | 0.777 | 0.754 | 0.461 | | 83.59 | 50 | 87.5 |
| Synthetic4 | k-means | 6 | 6 | 0.816 | 0.941 | 0.82 | 0.407 | 0.72 | 51.27 | 0 | 0 |
| | k-means++ | | | 0.958 | 0.988 | 0.932 | 0.222 | 0.62 | 10.96 | 0 | 92.63 |
| | fuzk | | | 0.98 | 0.994 | 0.953 | 0.183 | 0.45 | 8.738 | 0 | 94.5 |
| | AMOC(rand) | 28 | 2.875 | 0.444 | 0.719 | 0.696 | 0.6 | 0.682 | 88.28 | 87.5 | 100 |
| | AMOC(kmpp) | | 5.7 | 0.969 | 0.991 | 0.953 | 0.188 | 0.244 | 25.31 | 0 | 100 |
| | ACDE | | 7.9 | 0.979 | 0.994 | 0.878 | 0.308 | 0.359 | 53.21 | 0 | 93.88 |
| Synthetic5 | k-means | 3 | 3 | 0.197 | 0.62 | 0.396 | 1.176 | 1.78 | 53.9 | 51.67 | 56.11 |
| | k-means++ | | | 0.201 | 0.622 | 0.398 | 1.133 | 1.678 | 54.42 | 51.67 | 56.11 |
| | fuzk | | | 0.256 | 0.65 | 0.369 | 1.301 | 4.34 | 48.61 | 46.67 | 48.89 |
| | AMOC(rand) | 14 | 2 | 0.267 | 0.633 | 0.515 | 1.102 | 1.873 | 69.94 | 69.44 | 70 |
| | AMOC(kmpp) | | 2.3 | 0.244 | 0.627 | 0.482 | 1.118 | 1.854 | 65.22 | 45 | 70 |
| | ACDE | | 4.4 | 0.596 | 0.805 | 0.074 | 1.453 | 4.061 | 71.31 | 17.78 | 92.22 |
| Iris | k-means | 3 | 3 | 0.774 | 0.892 | 0.804 | 0.463 | 0.607 | 15.77 | 4 | 51.33 |
| | k-means++ | | | 0.796 | 0.904 | 0.804 | 0.461 | 0.712 | 13.37 | 4 | 51.33 |
| | fuzk | | | 0.788 | 0.899 | 0.803 | 0.46 | 0.658 | 15.33 | 4 | 56 |
| | AMOC(rand) | 12 | 2.133 | 0.61 | 0.799 | 0.932 | 0.259 | 0.429 | 29.42 | 4 | 33.33 |
| | AMOC(kmpp) | | 2.533 | 0.737 | 0.869 | 0.874 | 0.337 | 0.512 | 17.69 | 4 | 33.33 |
| | ACDE | | 3.15 | 0.887 | 0.95 | 0.784 | 0.435 | 0.706 | 10.17 | 3.333 | 62.67 |
| Wine | k-means | 3 | 3 | 0.295 | 0.675 | 0.694 | 0.569 | 0.612 | 34.58 | 30.34 | 42.7 |
| | k-means++ | | | 0.305 | 0.681 | 0.694 | 0.562 | 0.678 | 33.54 | 30.34 | 42.7 |
| | fuzk | | | 0.34 | 0.7 | 0.696 | 0.566 | 0.753 | 30.34 | 29.78 | 30.9 |
| | AMOC(rand) | 13 | 2 | 0.197 | 0.593 | 0.714 | 0.644 | 1.025 | 41.01 | 30.34 | 41.01 |
| | AMOC(kmpp) | | 2 | 0.197 | 0.593 | 0.714 | 0.644 | 1.025 | 41.01 | 41.01 | 41.01 |
| | ACDE | | 4.45 | 0.367 | 0.723 | 0.373 | 0.555 | 1.626 | 52.89 | 41.01 | 69.66 |
| Glass | k-means | 6 | 6 | 0.245 | 0.691 | 0.507 | 0.901 | 0.967 | 55.86 | 28.65 | 67.29 |
| | k-means++ | | | 0.259 | 0.683 | 0.548 | 0.871 | 1.523 | 56.1 | 44.86 | 64.95 |
| | fuzk | | | 0.241 | 0.72 | 0.293 | 0.998 | 1.613 | 62.29 | 46.73 | 66.82 |
| | AMOC(rand) | 15 | 3.333 | 0.231 | 0.618 | 0.618 | 0.96 | 1.808 | 68.75 | 48.13 | 76.64 |
| | AMOC(kmpp) | | 4.067 | 0.25 | 0.635 | 0.655 | 0.816 | 1.414 | 66.42 | 51.21 | 76.17 |
| | ACDE | | 5.5 | 0.309 | 0.712 | 0.338 | 1.146 | 2.868 | 54.35 | 57.48 | 86.45 |
| Yeast1 | k-means | 4 | 4 | 0.497 | 0.765 | 0.466 | 1.5 | 1.439 | 35.74 | 37.38 | 80.17 |
| | k-means++ | | | 0.465 | 0.751 | 0.425 | 1.528 | 1.678 | 37.49 | 35.02 | 42.62 |
| | fuzk | | | 0.43 | 0.734 | 0.37 | 2.012 | 1.679 | 39.18 | 35.02 | 80.59 |
| | AMOC(rand) | 15 | 3 | 0.476 | 0.749 | 0.443 | 1.558 | 1.609 | 79.35 | 37.55 | 80.59 |
| | AMOC(kmpp) | | 4.867 | 0.471 | 0.749 | 0.429 | 1.542 | 1.643 | 37.25 | 38.06 | 80.59 |

| | | | | | | | | | | |
|---|---|---|---|---|---|---|---|---|---|---|
| | ACDE | | 5.55 | 0.594 | 0.806 | 0.348 | 2.314 | 2.669 | 81.86 | 35.44 | 97.47 |
| Yeast2 | k-means | 5 | 5 | 0.447 | 0.803 | 0.438 | 1.307 | 1.721 | 38.35 | 24.47 | 57.03 |
| | k-means++ | | | 0.436 | 0.801 | 0.421 | 1.292 | 1.521 | 40 | 27.08 | 57.03 |
| | fuzk | | | 0.421 | 0.799 | 0.379 | 1.443 | 1.341 | 35.73 | 26.3 | 53.65 |
| | AMOC(rand) | 20 | 3.667 | 0.458 | 0.788 | 0.501 | 1.148 | 1.349 | 55.14 | 27.86 | 85.16 |
| | AMOC(kmpp) | | 4.4 | 0.476 | 0.805 | 0.492 | 1.155 | 1.391 | 38.21 | 26.56 | 44.53 |
| | ACDE | | 6.225 | 0.537 | 0.838 | 0.363 | 1.438 | 2.326 | 44.95 | 23.18 | 86.46 |

Figure1. Number of clusters for Yeast2

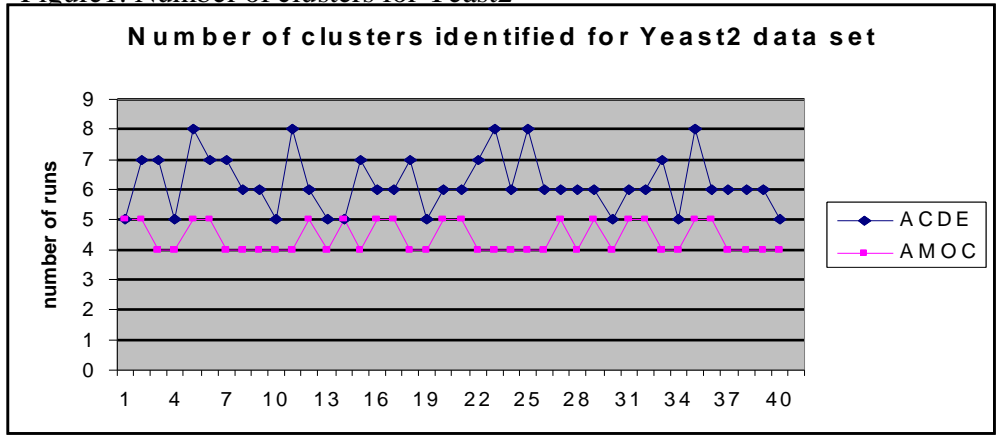

Figure2. Number of clusters of Synthetic2 data set

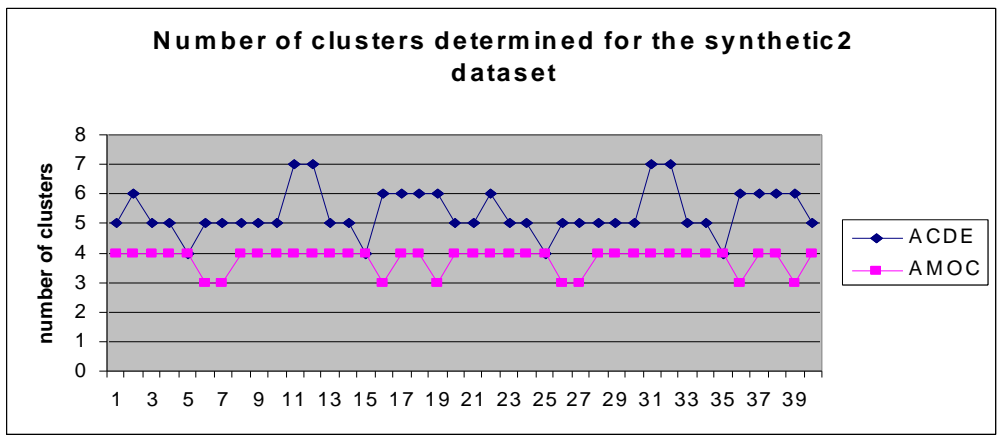

Figure3. Error rates obtained for Iris data set

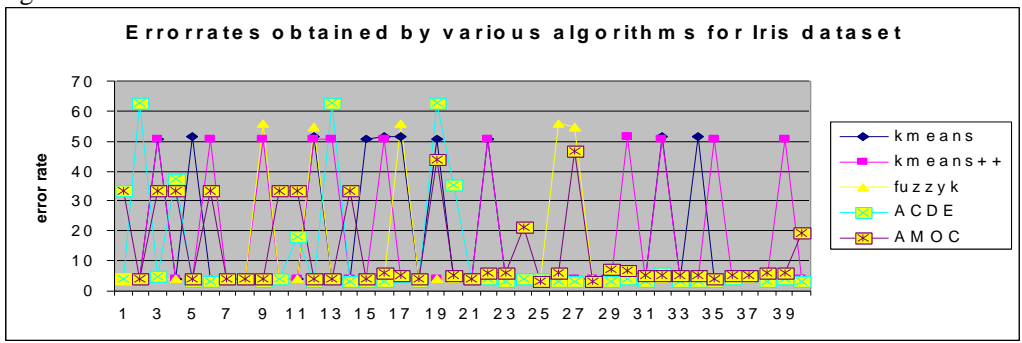

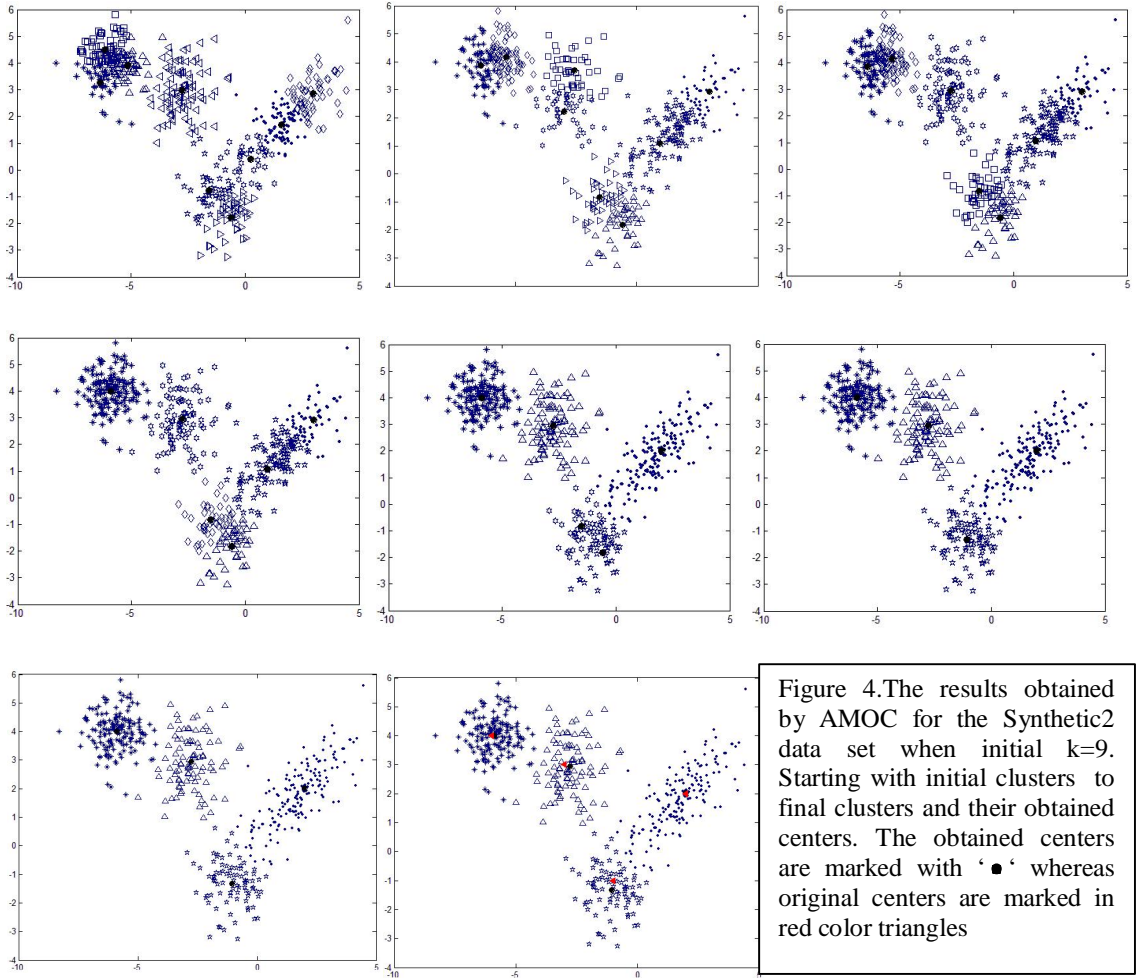

Figure 4.The results obtained by AMOC for the Synthetic2 data set when initial k=9. Starting with initial clusters to final clusters and their obtained centers. The obtained centers are marked with '●' whereas original centers are marked in red color triangles

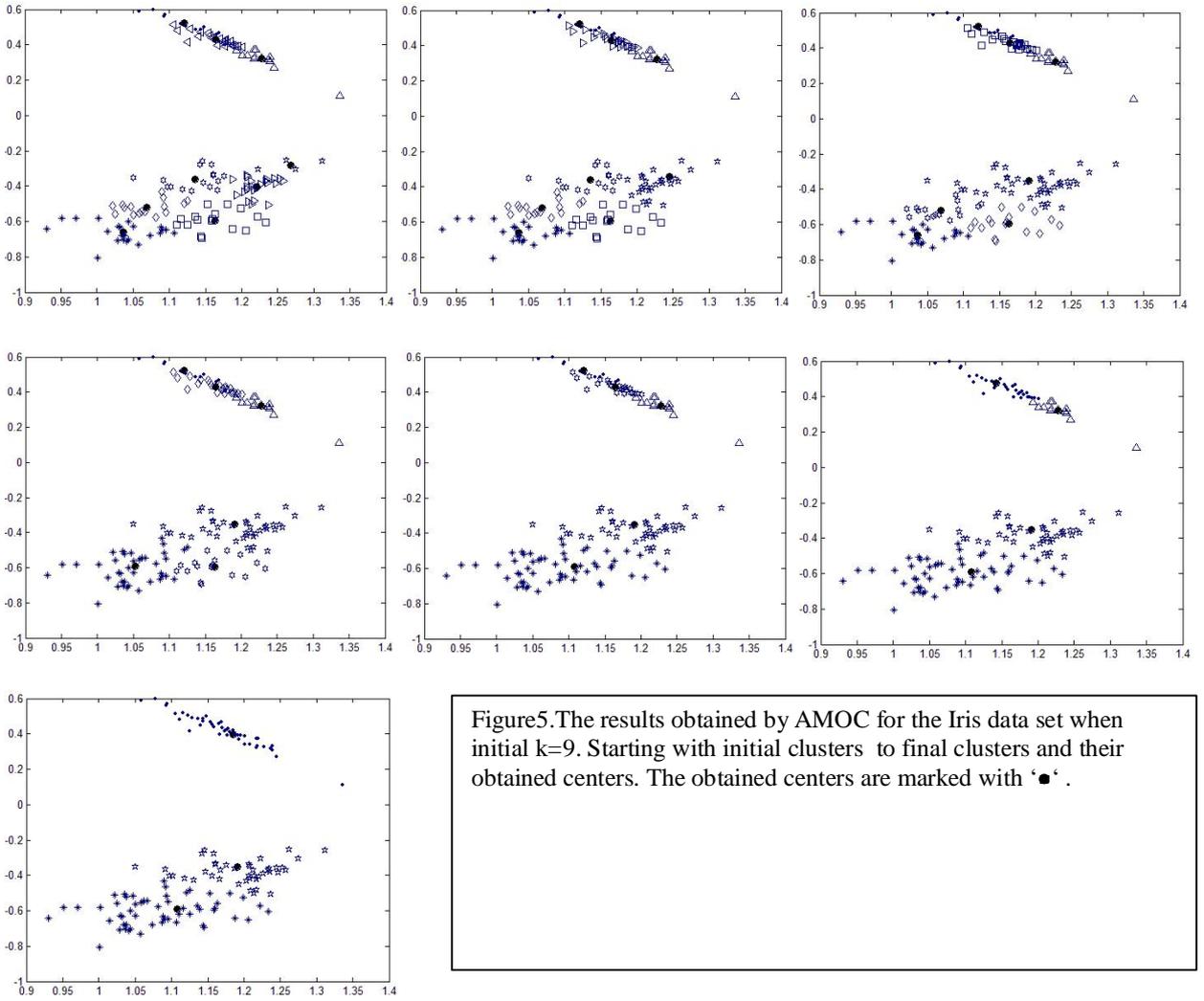

Figure 5. The results obtained by AMOC for the Iris data set when initial k=9. Starting with initial clusters to final clusters and their obtained centers. The obtained centers are marked with '●'.

Table 2. error rates of various algorithms

| Data set | AMOC(rand) | AMOC(kmpp) | SPSS | k-means | kmpp | Fuzzy-k | ACDE |
|---|---|---|---|---|---|---|---|
| synthetic1 | 2.209 | 1.905 | 1.714 | 2.236 | 1.914 | 2.571 | 51.56 |
| synthetic2 | 46.5 | 8.79 | 2.4 | 19.1 | 7.16 | 2.2 | 58.89 |
| Synthetic3 | 7.6 | 4.317 | 1 | 2.242 | 1 | 1 | 83.59 |
| Synthetic4 | 88.28 | 25.31 | 0 | 51.27 | 10.96 | 8.738 | 53.21 |
| synthetic5 | 69.94 | 65.22 | 52.22 | 53.9 | 54.42 | 48.61 | 71.31 |
| Iris | 29.42 | 17.69 | 50.67 | 15.77 | 13.37 | 15.33 | 10.17 |
| Wine | 41.01 | 41.01 | 30.34 | 34.58 | 33.54 | 30.34 | 52.89 |
| Glass | 68.75 | 66.42 | 45.79 | 55.86 | 56.1 | 62.29 | 54.35 |
| Yeast1 | 79.35 | 37.25 | 35.44 | 35.74 | 37.49 | 39.18 | 81.86 |
| Yeast2 | 55.14 | 38.21 | 43.23 | 38.35 | 40 | 35.73 | 44.95 |

**Comments on the results of AMOC**

The errors rates obtained from various algorithms vs data are presented in table 2.

- From the above table it is observed that the AMOC either producing best clusters than ACDE or performing equally well
- The results of AMOC show that average error rates is equally good when compared to those of k-means, k-means++, fuzzyk and SPSS
- The results of AMOC show that they are far better when compared to ACDE in most of the case.
- The best error rate over 40 runs of AMOC is very much comparable to the existing algorithms mentioned in the above observations.
- The maximum error rate over 40 runs of AMOC appears to be the least when compared to those of existing algorithms.
- The quality of AMOC in terms of Rand index is 70%.
- **Recently Sudhakar Jonnalagadda and Rajagopalan Srinivasan [17] developed a method that determined 5 clusters from yeast2 data set where as the almost all existing methods finds as 4. The proposed AMOC is also find 5 clusters from yeast2 data**

**Note**: Results of CS, HI, ARI, etc., are very much in agreement with above all observations in the performance of AMOC, hence detailed note with respect to them is not provided to avoid duplication.

## 5. Conclusion

AMOC is ideally free from parameter. Though the AMOC require possible large k as input, the input number of clusters does not affect the output number of clusters. The experimental results have shown the performance of AMOC in finding optimal clusters automatically.